\documentclass[letterpaper, 10 pt, conference]{ieeeconf}  

\IEEEoverridecommandlockouts                              

\usepackage{graphics}
\usepackage{epsfig}
\usepackage{amsmath, amsthm, amssymb}
\usepackage{subcaption}
\usepackage{xcolor}
\usepackage[bookmarks=true]{hyperref}
\usepackage[font=small]{caption}
\usepackage{textcomp}
\usepackage{makecell}
\usepackage{colortbl}
\usepackage{subcaption}
\usepackage{bbm}

\definecolor{mypink1}{rgb}{0.929, 0.490, 0.192}
\definecolor{mygreen1}{rgb}{0.662, 0.819, 0.556}
\definecolor{aliceblue}{rgb}{0.94, 0.97, 1.0}
\definecolor{light-gray}{rgb}{0.95,0.95,0.95}
\definecolor{LightCyan}{rgb}{0.776470,0.843137,0.933333}
\title{\LARGE \bf
Robot Tactile Gesture Recognition Based on Full-body Modular E-skin}

\author{Shuo Jiang$^{1}$, Boce Hu$^{1}$, Linfeng Zhao$^{1}$ and Lawson L.S. Wong$^{1}$
\thanks{$^{1}$ The authors are with Northeastern University, Boston, MA, USA
        {\tt\small $\{$jiang.shuo, hu.boce, zhao.linf, l.wong$\}$@northeastern.edu}}%
}

\begin{document}

\maketitle
\thispagestyle{empty}
\pagestyle{empty}

\begin{abstract}

With the development of robot electronic skin technology, various tactile sensors, enhanced by AI, are unlocking a new dimension of perception for robots. In this work, we explore how robots equipped with electronic skin can recognize tactile gestures and interpret them as human commands. We developed a modular robot E-skin, composed of multiple irregularly shaped skin patches, which can be assembled to cover the robot’s body while capturing real-time pressure and pose data from thousands of sensing points. To process this information, we propose an equivariant graph neural network-based recognizer that efficiently and accurately classifies diverse tactile gestures, including poke, grab, stroke, and double-pat. By mapping the recognized gestures to predefined robot actions, we enable intuitive human-robot interaction purely through tactile input.

\end{abstract}

\section{INTRODUCTION}
With the advancements in AI and robotics, various service robots are expected to become integrated into human society in the foreseeable future, collaborating closely with humans. In response to the increasing need for human-robot interaction (HRI), tactile sensing, as an emerging technology in robotics, is anticipated to significantly enhance HRI performance in aspects such as obstacle avoidance, grasping, and instruction interpretation. Research on tactile interactions among humans \cite{yohanan2012role} and between humans and animals \cite{rew2000friends, vormbrock1988cardiovascular} indicates that humans often employ specific gestures that involve tactile actions, such as hugging, double-patting, and stroking, to express emotions or convey instructions. These gestures, commonly referred to as tactile gestures, can provide robots with the ability to better comprehend multimodal human commands and produce more comforting interactive responses.

In the context of robotics, the recognition of tactile gestures is based on large-area high-resolution electronic skin and is completed through the following subsequent processes. First, spatio-temporal signals, typically normal force signals, are collected from a large-scale tactile sensor array. These signals are then processed using machine learning algorithms to train classifiers, and the classification results are subsequently mapped to the corresponding robot actions. Robot tactile gesture recognition presents several challenges: (1) tactile gestures inherently comprise spatiotemporal signals, resulting in large data volume and high data dimensionality; (2) recognizing tactile gestures necessitates the use of large-scale, high-resolution electronic skins with multicontact recognition capability, whose fabricating process is notably intricate; (3) skins attached to robots may deform with the robot's movements, rendering the spatial composition of tactile signals inconsistent. For example, when you hold an apple with both hands, the skin of your two hands temporarily forms a continuous perception surface. However, the skin on the two hands is actually two separate regions on the whole skin surface of the body.
\begin{figure}[!t]
 \centering
 \includegraphics[width=.9\linewidth]{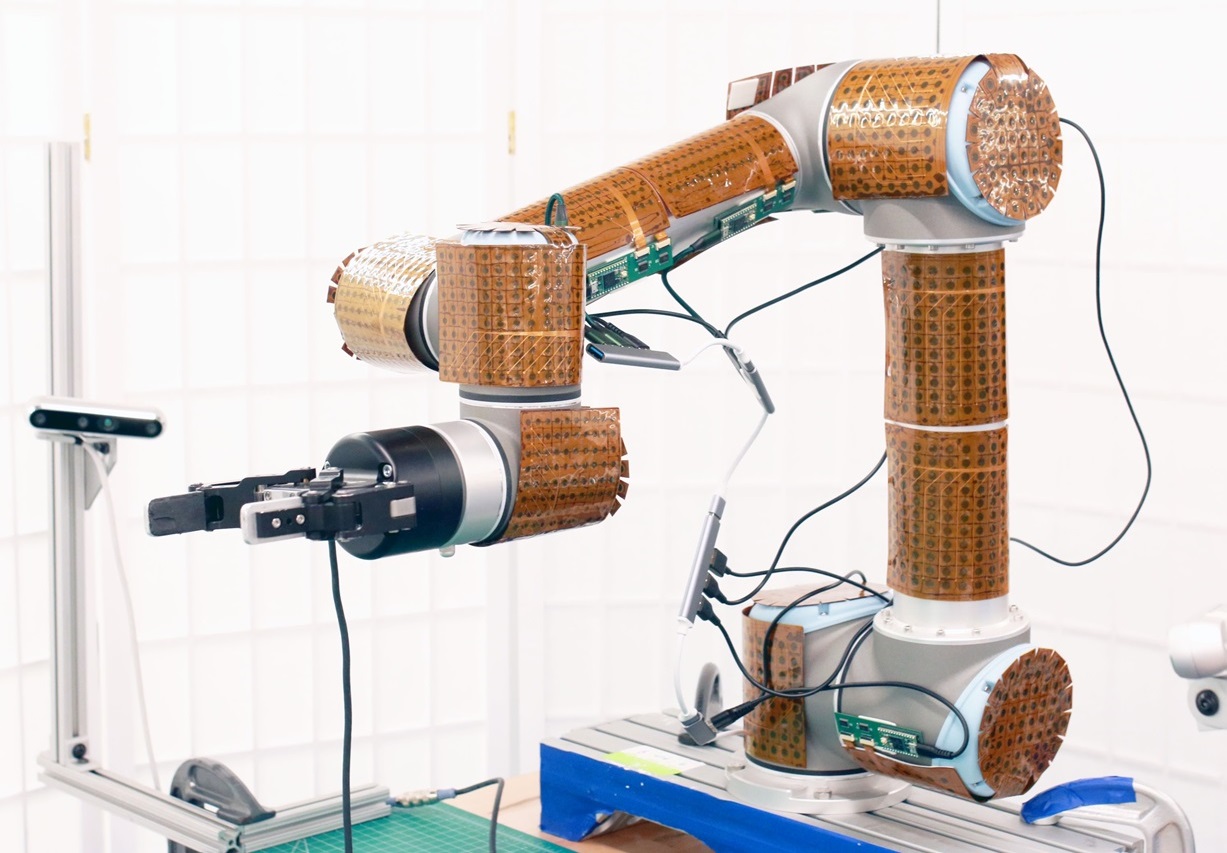}
 \caption{Our modular robot skin system enables customized robot body coverage and high-speed tactile gesture recognition.}
 \label{fig_21}
\vspace{-10pt}
\end{figure}

Most existing solutions for tactile gesture recognition utilize rectangular skin patches attached to simple geometric surfaces, such as flat or cylindrical shapes \cite{lin2021event, choi2022deep, sarwar2021large}. Although these solutions can achieve high accuracy in gesture recognition, their inability to transfer effectively to robot surfaces with different and time-varying geometry poses a significant limitation. For instance, solutions based on convolutional neural networks (CNNs) assume that the relative spatial positions of all sensing points remain constant, a condition only met when a single skin patch is used and attached to a rigid flat surface. 
In addition, due to the absence of a unified manufacturing standard, resolutions, sizes, and shapes of different skin solutions vary significantly, resulting in difficulty in model transferability.
\begin{table*}[!t]
    \centering
    \begin{tabular}{|c | c c c c c c c c|} 
 \hline
  \rowcolor{LightCyan}& Num of taxels & Density & Frequency & Mechanism & Num of gestures & Robot & Classifiers & Accuracy \\ [0.55ex] 
 \hline\hline
 \rowcolor{light-gray}
     \cellcolor[rgb]{0.772549, 0.878431, 0.705882}\cite{lin2021event} & 672 & 0.64/$\mathrm{cm}^{2}$ & 8Hz & Capacitive & 6 & Humanoid & SVM, KNN & 68 \% \\ 
 \hline
 \rowcolor{light-gray}
     \cellcolor[rgb]{0.772549, 0.878431, 0.705882}\cite{jung2017automatic} & 64 & 0.25/$\mathrm{cm}^{2}$ & 20Hz & Piezoresistive & 14 & Forearm & SVM, MLP & 60 \% \\ 
 \hline
 \rowcolor{light-gray}
     \cellcolor[rgb]{0.772549, 0.878431, 0.705882}\cite{kong2022bioinspired} & 300 & 0.92/$\mathrm{cm}^{2}$ & 50Hz & Piezoresistive & 5 & Flat & MLP & 76 \% \\ 
 \hline
 \rowcolor{light-gray}
     \cellcolor[rgb]{0.772549, 0.878431, 0.705882}\cite{sun2017categories} & 160 & 1.56/$\mathrm{cm}^{2}$ & 30Hz & Piezoresistive & 7 & Upperarm & KNN & 71 \% \\ 
 \hline
 \rowcolor{light-gray}
     \cellcolor[rgb]{0.772549, 0.878431, 0.705882}\cite{choi2022deep} & 45 & 0.44/$\mathrm{cm}^{2}$ & 25Hz & Magnetic & 13 & Forearm & CNN & 74 \% \\ 
 \hline
 \rowcolor{light-gray}
     \cellcolor[rgb]{0.772549, 0.878431, 0.705882}\cite{alonso2017detecting} & 1 & - & 172Hz & Acoustic & 4 & Head top & CNN, SVM, MLP & 75 \% \\ 
 \hline
 \rowcolor{light-gray}
     \cellcolor[rgb]{0.772549, 0.878431, 0.705882}\cite{zhan2023enable} & 6772 & 4.0/$\mathrm{cm}^{2}$ & 10Hz & Piezoresistive & 81 & Quadrupedal & CNN & 98.7 \% \\ 
 \hline
 \rowcolor{light-gray}
     \cellcolor[rgb]{0.772549, 0.878431, 0.705882}\cite{sarwar2021large} & 160 & 0.61/$\mathrm{cm}^{2}$ & 8Hz & Capacitive & 6 & Forearm & 3DCNN & 99 \% \\ 
 \hline
 \rowcolor{light-gray}
     \cellcolor[rgb]{0.772549, 0.878431, 0.705882}\cite{wang2021organization} & 1024 & 2.8/$\mathrm{cm}^{2}$ & 100Hz & Capacitive & 12 & Flat & CNN & 90 \% \\ 
 \hline
 \rowcolor{light-gray}
     \cellcolor[rgb]{0.772549, 0.878431, 0.705882}\cite{kubus2017robust} & 2048 & 4.0/$\mathrm{cm}^{2}$ & 42Hz & Piezoresistive & 16 & Upperarm & SVM & 97 \% \\ 
 \hline
 \rowcolor{light-gray}
     \cellcolor[rgb]{0.772549, 0.878431, 0.705882}\cite{zhou2017textile} & 400 & 1.0/$\mathrm{cm}^{2}$ & 50Hz & Piezoresistive & 7 & Forearm & SVM, KNN, LDA & 89.1 \% \\ 
 \hline
\rowcolor{light-gray}
     \cellcolor[rgb]{0.772549, 0.878431, 0.705882}Ours & 2112 & 1.56/$\mathrm{cm}^{2}$ & 50Hz & Piezoresistive & 5 & Robot arm & GNN &  96.1\% \\ 
 \hline
\end{tabular}
    \caption{State of the art comparison}
    \label{tab_1}
\end{table*}

To address the aforementioned challenges, we first developed a modular robot e-skin based on piezoresistance and flexible printed circuit technology. By fabricating skin modules with irregular shapes and customizing the interconnection among them, we can theoretically adapt these skins to various robot forms. We have practically tested this system on a robot arm. Subsequently, we proposed a tactile gesture recognizer based on geometric deep learning \cite{bronstein2021geometric}. This approach is characterized by its versatility, as it is (1) insensitive to the size, shape, and density of the skin, and (2) unaffected by geometric changes due to the robot's movements. These features mark a significant departure from state-of-the-art tactile gesture recognition methods. Furthermore, we deployed our skin system and tactile gesture recognition solution in a human-robot collaboration task, validating that operators can fully control the robot through tactile interactions alone, establishing a novel approach to human-robot interaction.

\begin{figure}
\begin{subfigure}{.5\linewidth}
    \centering
    \includegraphics[height=.12\textheight]{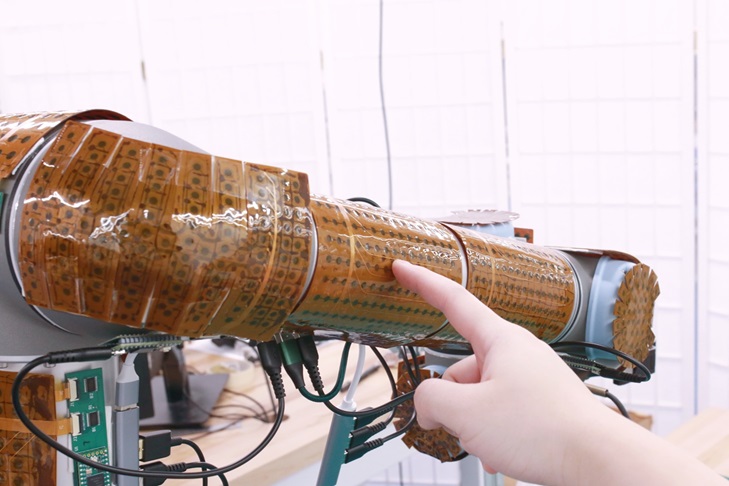}
    \caption{}
\end{subfigure}
\begin{subfigure}{.48\linewidth}
    \centering
    \includegraphics[height=.12\textheight]{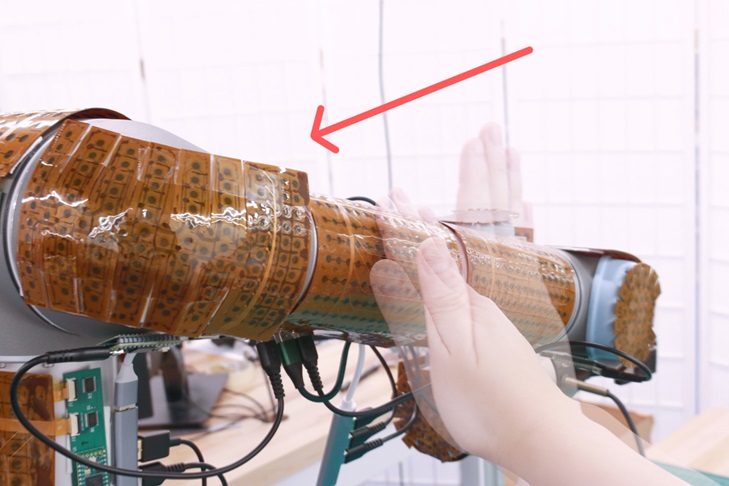}
    \caption{}
\end{subfigure}
\begin{subfigure}{.5\linewidth}
    \centering
    \includegraphics[height=.12\textheight]{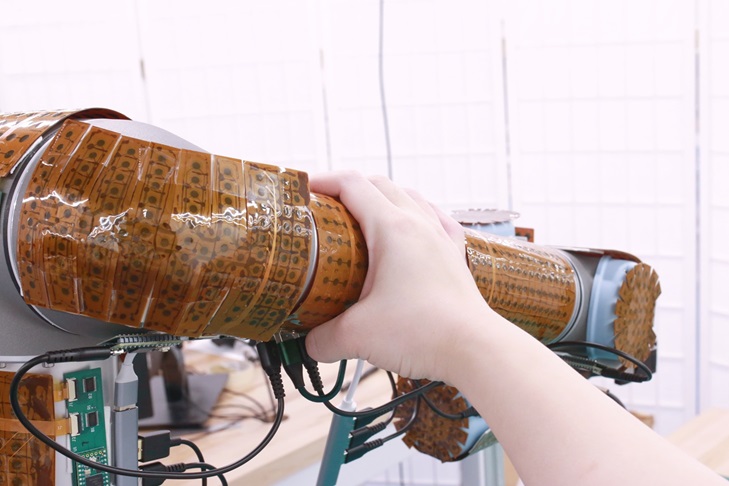}
    \caption{}
\end{subfigure}
\begin{subfigure}{.48\linewidth}
    \centering
    \includegraphics[height=.12\textheight]{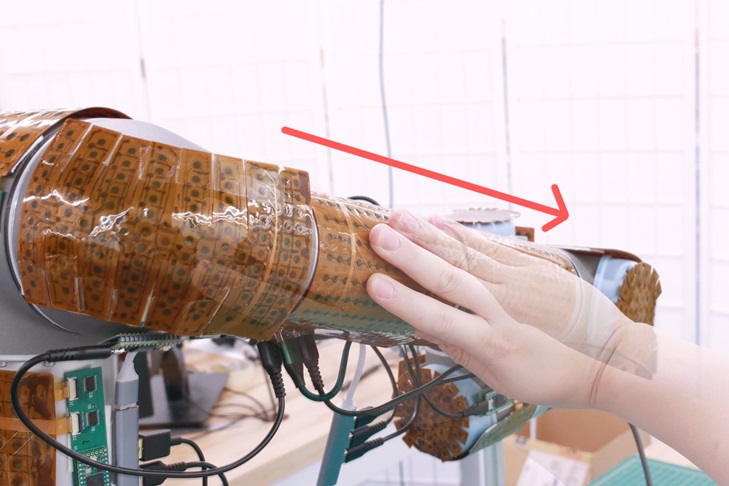}
    \caption{}
\end{subfigure}
\caption{Interactive tactile gestures investigated in this work: (a) poke; (b) double-pat; (c) grab; (d) stroke.}
\label{fig_31}
\vspace{-10pt}
\end{figure}
\section{RELATED WORKS}
We summarize some representative tactile gesture recognition approaches in Table \ref{tab_1}, comparing the characteristics of their skin units and classifier performance. These methods employ various skin sensor technologies, such as capacitive, piezoresistive, magnetic, and acoustic, for tactile gesture recognition. Although many previous methods have achieved very high recognition accuracy, this does not mean the problem is fully solved. We believe there is still room for improvement in the following aspects: recognition speed—while many approaches attain high accuracy, they often suffer from low computational efficiency, making real-time deployment on robots challenging; tactile skin structure—most existing tactile skin systems consist of uniformly distributed taxels with regular shapes and simple, flat, or mildly curved surfaces. This significantly simplifies the problem but does not fully address more complex, real-world applications. The primary challenge of tactile gesture recognition is efficiently processing sensor signals with spatiotemporal characteristics. Traditional machine learning approaches usually employ window-based techniques to process temporal signals and use specific feature extraction methods \cite{lin2021event, sun2017categories} or spline fitting \cite{kubus2017robust} for spatial feature extraction. On the other hand, deep learning-based methods primarily utilize network architectures with geometric recognition capabilities, such as CNNs \cite{zhan2023enable, albini2020pressure, albini2018tactile} (which exhibit translation equivariance) or graph neural networks (GNNs) \cite{garcia2019tactilegcn, funabashi2022multi, yang2023tacgnn} to process spatial information, while recurrent layers are used for temporal analysis \cite{sarwar2021large}. However, a limitation of most CNN-based works is their reliance on rectangular skin shapes with uniformly arranged sensor points. In \cite{albini2020pressure}, a solution was proposed for processing complex non-planar 2D skin signals using CNNs. They introduced the concept of a tactile image, which is obtained by resampling data from flattening the 3D mesh of the tactile sensor grid, effectively transforming the problem into an image processing task. However, this approach has several limitations. First, flattening a curved surface inevitably introduces image distortion. Second, it is challenging to dynamically extend the method to multiple skin patches. In \cite{albini2018tactile}, an improvement was made to handle multiple skin patches, but the approach is not sufficiently general—it can only connect inherently adjacent skin regions and fails to accommodate skin patches that are temporarily connected due to robot movement (Fig. \ref{fig_7} (bottom). Most GNN-based tactile signal processing approaches are limited to small-scale skin applications \cite{garcia2019tactilegcn, fan2022graph}, primarily designed to handle irregular taxel distributions. The geometric structure of the graph is less affected by robot movement, making these methods well-suited for tasks such as dexterous hand control \cite{yang2023tacgnn} and object recognition \cite{kulkarni2024tactile}. Therefore, the primary goal of this work is to bridge the gap in GNN-based solutions by addressing tactile gesture recognition and human-robot interaction on large-scale, highly dynamic tactile skins. One interesting work in \cite{doi:10.1126/scirobotics.adn4008} employed joint F/T sensors but not tactile sensors to estimate robot body tactile trajectory, and applied equivariant CNN to recognize the digits and English characters written on the robot body. However, the solutions has limitation that it can only recognize two simultaneous contacts and payload can significantly shift the reading of joint F/T sensors; these issues can be fully addressed in tactile sensing.
\section{PRELIMINARIES}
\subsection{$\mathrm{E}(n)$ Equivariance}
Equivariance is a form of symmetry for functions from one space with symmetry to another. Let $T_{g}:X\to X$ be a set of transformations on $X$ for the
abstract group $g\in G$. A function $\phi :X\to Y$ is equivariant to $g$ if there exists an equivalent transformation on its output space $S_{g}:Y\to Y$ such that:
\begin{equation}
    S_{g}\left ( \phi \left ( x \right ) \right )=\phi \left (  T_{g}\left (x\right ) \right )
\end{equation}
When a function is equivariant, it indicates its ability to faithfully preserve the symmetries of the original space through a mapping to the target space. Given that tactile gestures operate within the 3D Euclidean space $\mathrm{SE} \left ( 3 \right )$, they inherently possess symmetries within  $\mathrm{SE} \left ( 3 \right )$, and $T_{g}$ are homogeneous transformations in this context. This characteristic allows the model to effectively maintain the geometric structural properties of the data, enhance the model's generalization ability, and reduce the amount of training data required. It is extensively applied in the field of robotics \cite{doi:10.1126/scirobotics.adn4008, zhu2022sample}.
\begin{figure}[!t]
 \centering
 \includegraphics[width=\linewidth]{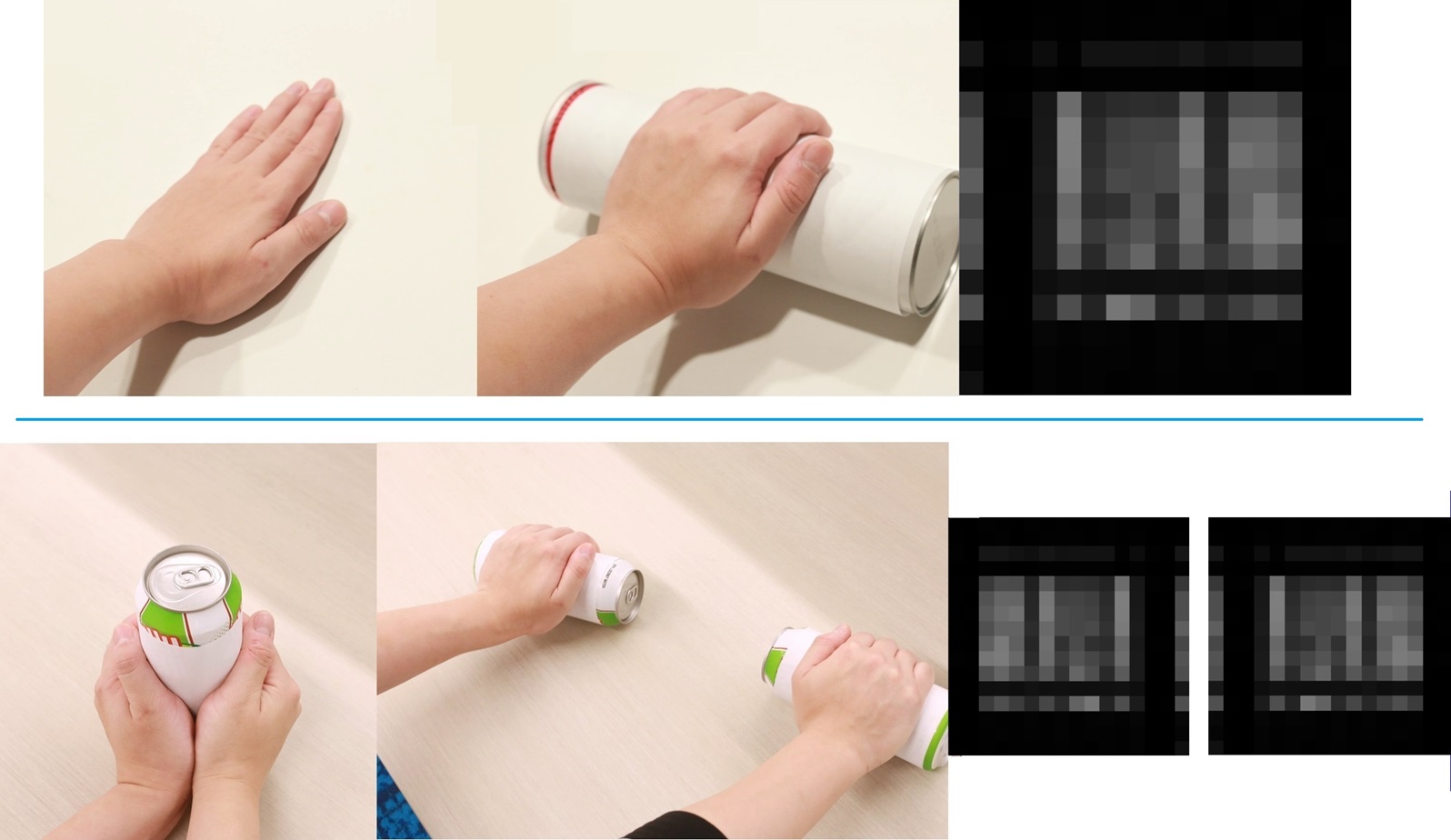}
 \caption{(top) ambiguity caused by deformation, in both cases the gestures generate the same pressure image; (bottom) ambiguity caused by temporary merging perceptual surfaces, in both cases the gestures generate the same pressure image.}
 \label{fig_7}
\vspace{-10pt}
\end{figure}
\subsection{Graph Neural Network}
Different from CNNs, GNNs are particularly designed to operate on graph-structured data, e.g., the irregularly arranged tactile sensor array. Given a graph $\mathcal{G}=\left ( \mathcal{V}, \mathcal{E} \right )$ with nodes $v_{i}\in \mathcal{V}$ and edges $e_{ij}\in \mathcal{E}$, the Equivariant Graph Convolutional Layer (EGCL) at layer $l$ is designed to process the data by:
\begin{equation}
    \begin{split}
        m_{ij}&=\phi _{e}\left ( v_{i}^{l},v_{j}^{l},\left\|x_{i}^{l}-x_{j}^{l} \right\|^{2},e_{ij} \right )\\
        x^{l+1}&=x^{l}+C\sum _{i\neq j}\left ( x_{i}^{l} -x_{j}^{l}\right )\phi _{x}\left ( m_{ij} \right )\\
        m_{i}&=\sum _{i\neq j}m_{ij}\\
        v_{i}^{l+1}&=\phi _{h}\left ( v_{i}^{l},m_{i} \right )
    \end{split}
\end{equation}
where $x_{i}^{l}$ are node $i$'s poses, $\phi _{e}$ and $\phi _{h}$ are the
edge and node operations. The operation is proven to be $\mathrm{SE}(3)$ equivariant \cite{satorras2021n}.
\section{TACTILE GESTURE RECOGNITION}
In this work, we primarily address the problem of how to perceive and recognize tactile gestures applied to a robot equipped with full-body electronic skin. We will introduce our robot skin hardware design and how we use a geometric deep learning-based classifier to recognize four types of tactile gestures, as shown in Fig. \ref{fig_31}: poke, double-pat, grab, and stroke. Our goal is to deploy the trained recognizers onto the robot, enabling humans to interact with the robot and issue commands through tactile gestures by mapping these gestures to robot actions.
\begin{figure}[!t]
\begin{subfigure}{.54\linewidth}
    \centering
    \includegraphics[height=.18\textheight]{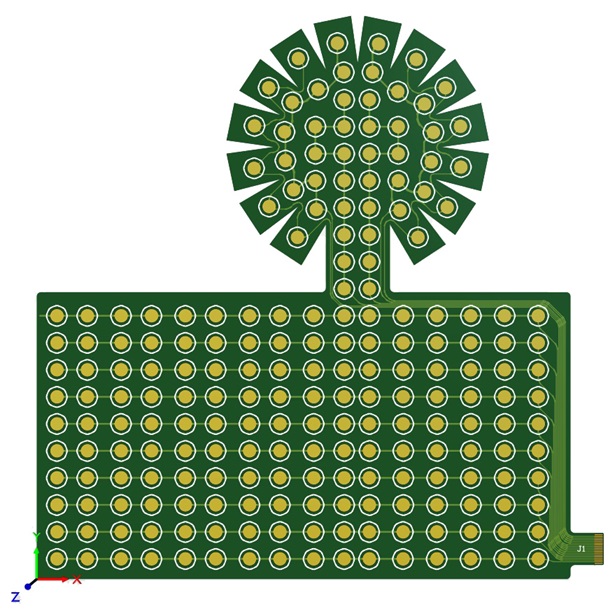}
    \caption{}
\end{subfigure}
\begin{subfigure}{.44\linewidth}
    \centering
    \includegraphics[height=.18\textheight]{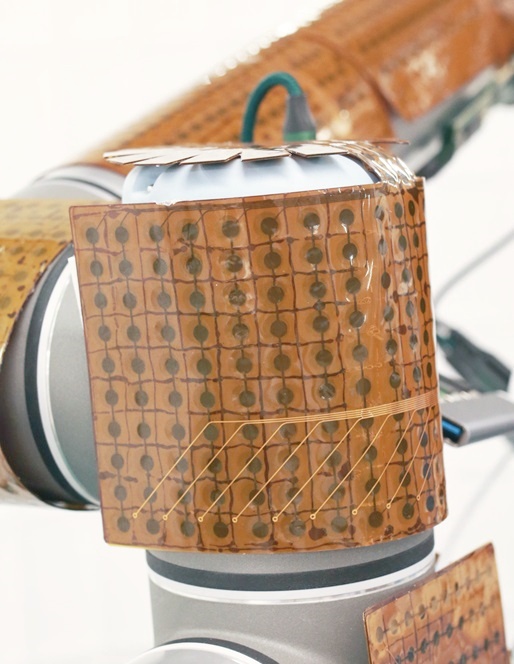}
    \caption{}
\end{subfigure}
 \caption{(a) irregular skin patch; (b) closer view at patch mounting.}
 \label{fig_9}
\vspace{-10pt}
\end{figure}
\begin{figure}[!t]
 \centering
\includegraphics[width=.5\linewidth]{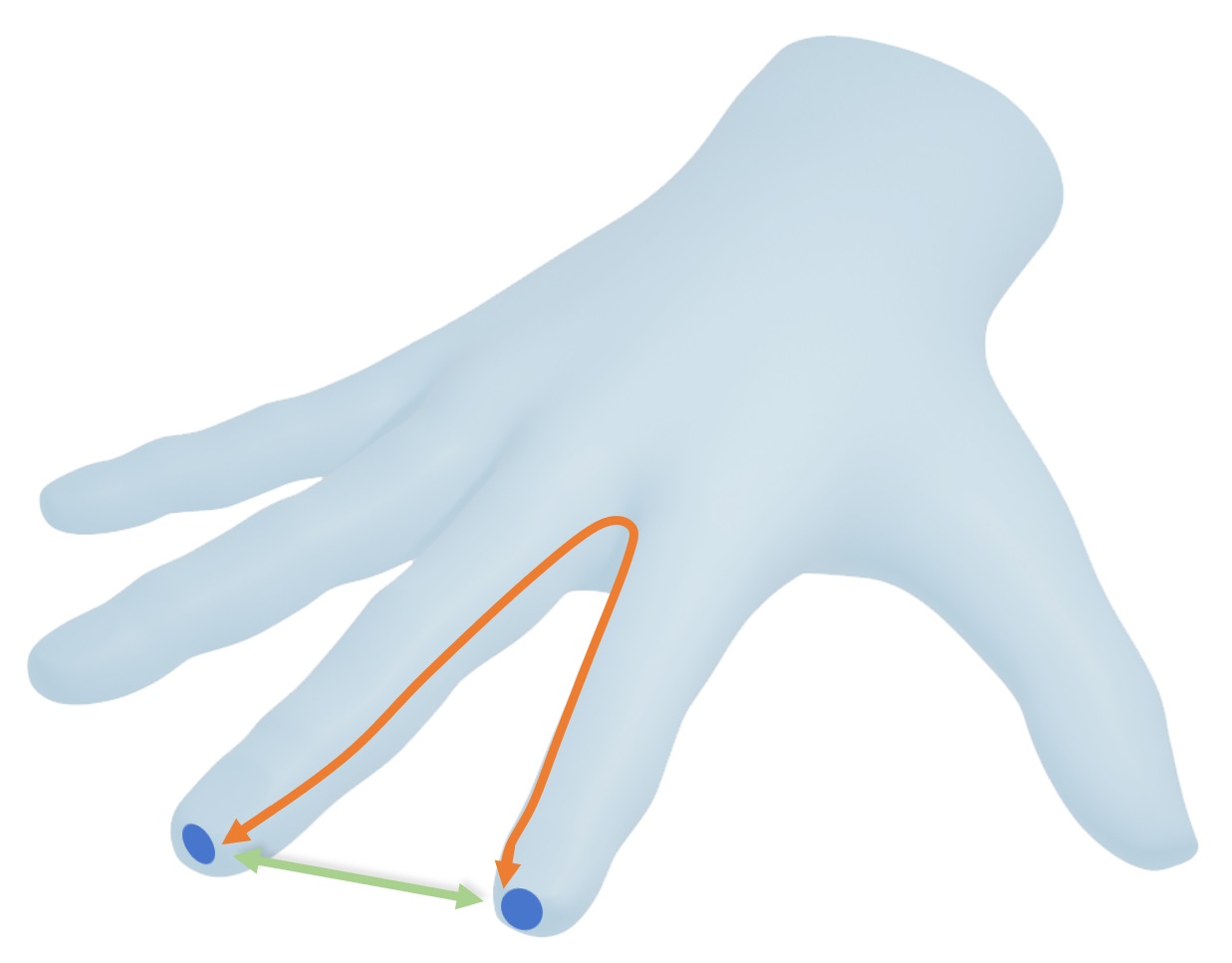}
 \caption{Example of \textcolor{mypink1}{geometric distance} and \textcolor{mygreen1}{kinematic distance}.}
 \label{fig_8}
\vspace{-10pt}
\end{figure}
\subsection{Discussion}
Before designing the classifier, it is essential to analyze the factors that can influence tactile gesture recognition. The absence of crucial features can significantly impact the classifier's performance. 
\begin{figure*}[!t]
\begin{subfigure}{.4\linewidth}
    \centering
\includegraphics[height=.18\textheight]{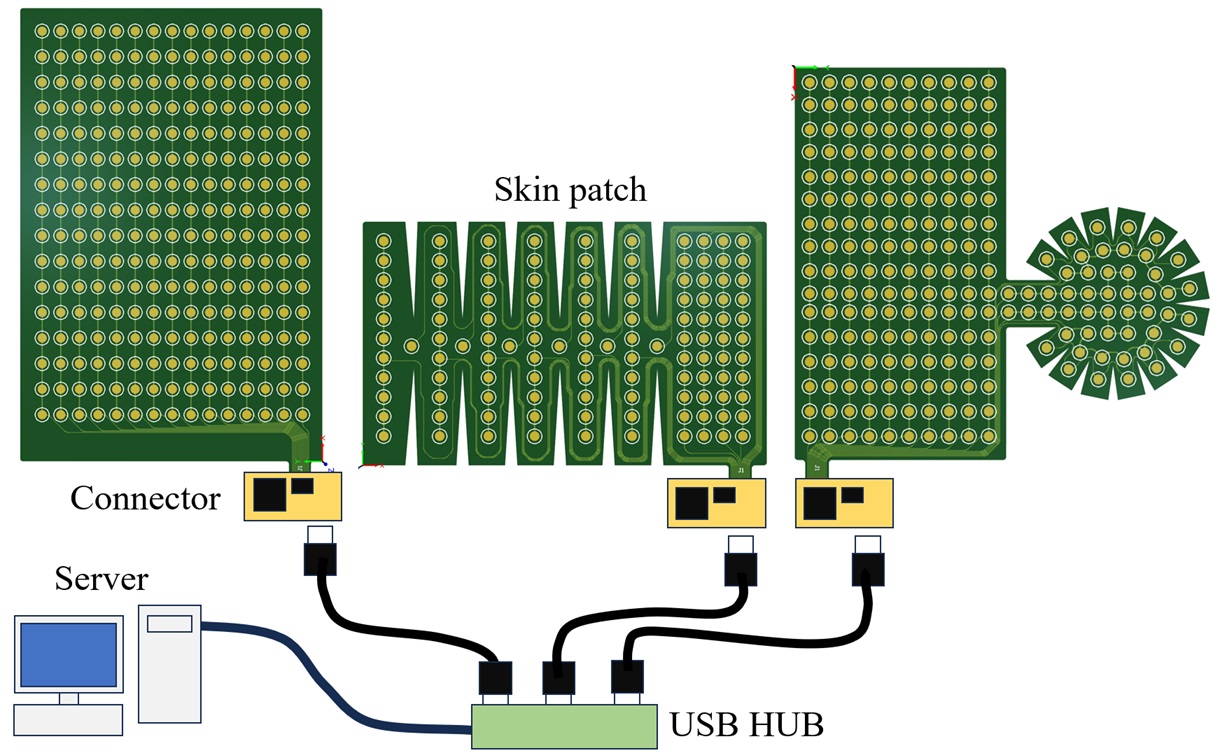}
    \caption{}
\end{subfigure}
\begin{subfigure}{.38\linewidth}
    \centering
\includegraphics[height=.18\textheight]{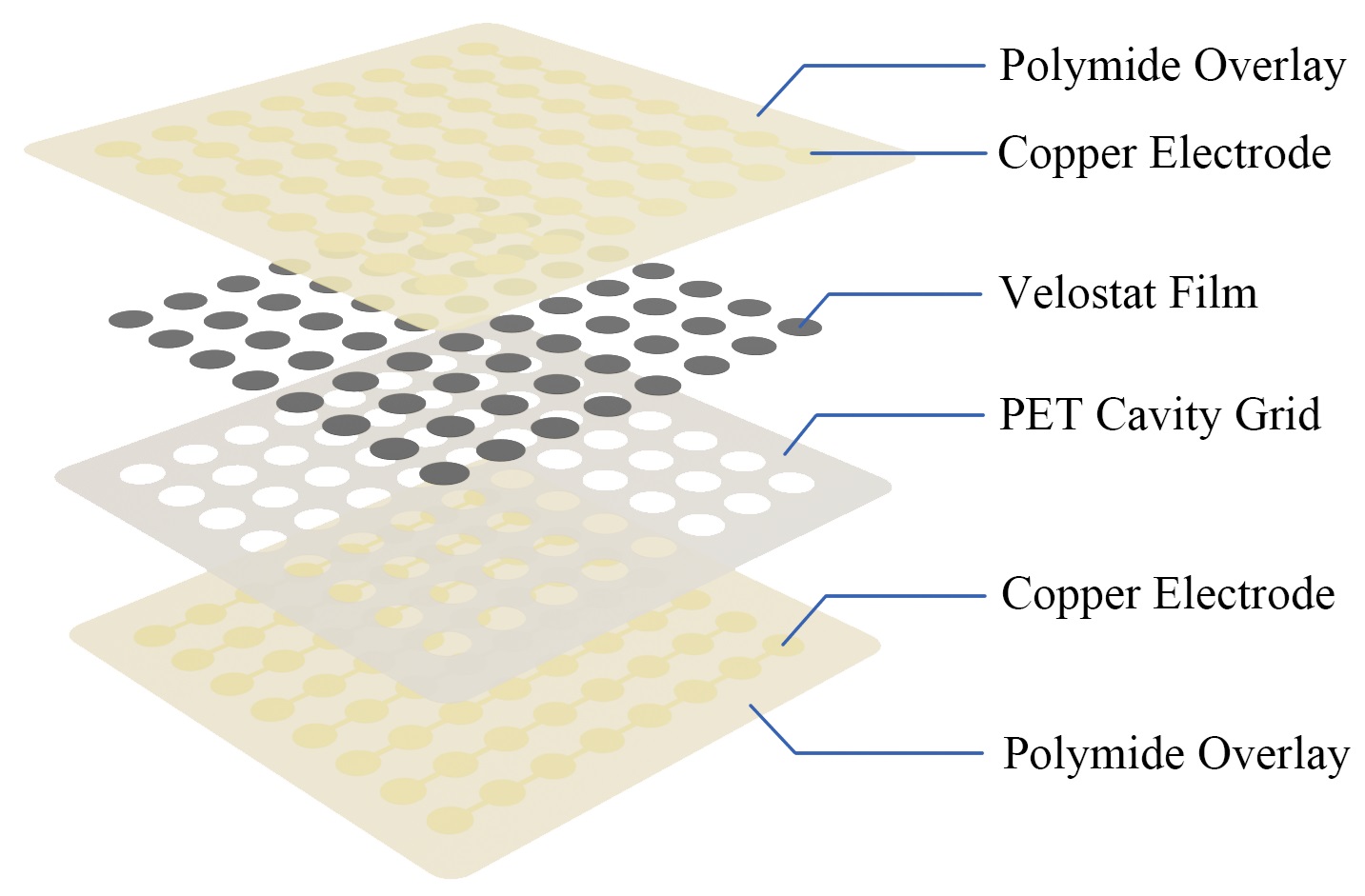}
    \caption{}
\end{subfigure}
\begin{subfigure}{.2\linewidth}
    \centering
\includegraphics[height=.18\textheight]{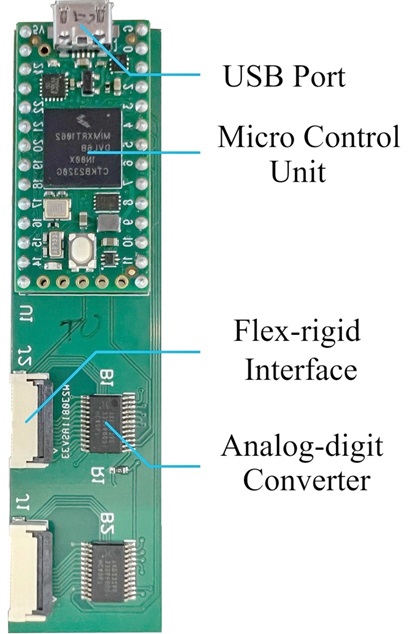}
    \caption{}
\end{subfigure}
 \caption{(a) Skin perception system; (b) flexible skin patch fabricating; (c) Connector design.}
 \label{fig_27}
\vspace{-10pt}
\end{figure*}
\subsubsection{Shape}
Due to the presence of numerous curved surfaces and irregular shapes on the robot's surface, when adapting the skin to the robot's surface, the shapes of most skin patches are considered irregular, as for the arrangement of sensors, with variable spacing. For example, in Fig. \ref{fig_9}, we demonstrate a skin patch adapted to the cylindrical joint of a robot. The classifier should be capable of handling spatial signals that are non-flat and non-uniformly distributed.
\subsubsection{Density}
Assuming that two skin patches have the same number of sensors and the sensors are arranged in the same way (e.g., in a matrix), they are identical if we neglect the geometric information. However, if the two skin patches have different sensor densities, resulting in different sizes, the same sensor readings may have entirely different physical interaction meanings.
\subsubsection{Deformation}
As illustrated in the top row of Fig. \ref{fig_7}, the tactile gestures "touching" and "grabbing" can generate nearly identical tactile perception results if we rely solely on flattened pressure distribution to process tactile signals. This suggests that classification is not solely dependent on surface pressure distribution; rather, the deformation pattern (geometric change) influenced by the robot's kinematics also plays a crucial role in distinguishing gestures.
\subsubsection{Temporary Merging}
Another scenario is when two skin patches on the body surface, which are not adjacent, can be temporarily combined into a complete perceptual surface by the robot's body movement, as shown in Fig. \ref{fig_7} bottom. The perceptual surfaces on both hands are spatially distant at the skin scale, but through body movements, they can temporarily combine to perceive the same object while still remaining separable to simultaneously perceive two different objects. In other words, the formation of perceptual surfaces is dynamic.
\subsubsection{Distance}
According to our analysis, finding an effective way to describe the geometric relationships between sensors is crucial for tactile gesture recognition. Hence, we propose distance concepts tailored to different application scenarios, the geometric and kinematic distances.
\begin{itemize}
    \item The \textcolor{mypink1}{geometric distance} between two sensors is defined as the geodesic length on the robot surface manifold.
    \item The \textcolor{mygreen1}{kinematic distance} between two sensors is defined as the Euclidean distance in Cartesian space.
\end{itemize}
From Fig. \ref{fig_8}, we clearly observe the distinction between the geometric distance and the kinematic distance. In most cases, the robot skin can bend but is difficult to stretch, which means the geometric distance between two sensors on the surface remains constant. However, the kinematic distance changes due to skin deformation or spatial movement induced by the movement of the robot. Considering the points discussed above, we believe that kinematic distance holds more importance for the classifier in practice, as the perceptual surface undergoes dynamic geometric and topological changes with the robot's movement. Our goal is to develop a classifier that effectively incorporates kinematic distances into its learning process.
\subsection{Hardware}
In this work, we designed a modular electronic skin system that allows customizable coverage of the entire robot body, as shown in Fig. \ref{fig_27}a. It consists of flexible skin patches and hardware interfaces (Fig. \ref{fig_27} c), such flex-rigid separation design with modular skin patches, allowing replacement of any localized damage and ensuring the high durability of the system. The readings from multiple skin patches are sent to the server via USB 2.0 protocol (max 480 Mbps speed). The flexible skin patches are fabricated by laminating polymide overlays, Polyethylene TErephthalate (PET) grid, and velostat film (Fig. \ref{fig_27}b). External mechanical compression induces resistance changes in the velostat film, which are then captured by electrodes on both sides of the polymide overlays. 

We custom-made a skin system for an UR5 robot, as shown in Fig.\ref{fig_21}. This system consists of 2112 sensing points that collect normal pressure information at a rate of 50Hz. We are capable of collecting data at a higher frequency, but limited be the recognizer's speed, we have aligned our sampling rate to match that of the recognizer. The sensing points on skin patches at different locations vary in size and have different distribution densities. The skin closer to the end-effector of the robot has a higher sensing density. The distribution of the sensing points on the robot is shown in Fig. \ref{fig_15}a, with different skin patches distinguished by different colors.
\subsection{Data Processing}
Based on our previous discussion, geometric graph neural networks emerge as an ideal classifier for processing dynamic spatio-temporal geometric information. Its advantage lies in the excellent scalability and the absence of specific requirements for the relative positional relationships between sensors, e.g. arranging in a rectangular grid. We structure our GNN-based recognizer as follows: 
\begin{figure}[!t]
\begin{subfigure}{.45\linewidth}
    \centering
    \includegraphics[height=.18\textheight]{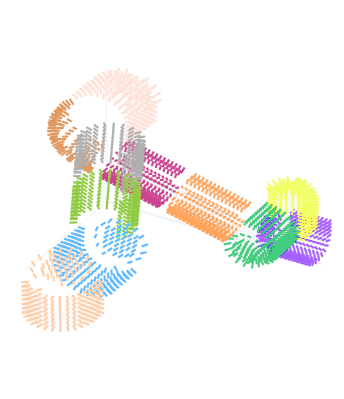}
    \caption{}
\end{subfigure}
\begin{subfigure}{.51\linewidth}
    \centering
    \includegraphics[height=.22\textheight]{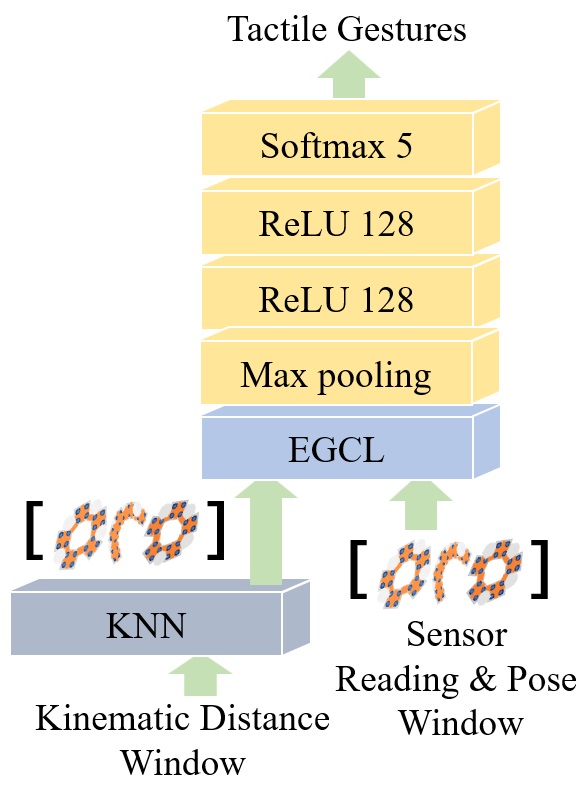}
    \caption{}
\end{subfigure}
 \caption{(a) Sensing points distribution. Different patches are grouped by colors; (b) neural network structure.}
 \label{fig_15}
\vspace{-10pt}
\end{figure}
\subsubsection{Node Feature}
In our approach, each sensor on the robot is associated with a specific node within the GNN, and the node's feature vector includes the scalar pressure reading at the corresponding sensing point. We assign each sensor a local coordinate frame, orienting the z-axis to align with the outward normal direction. The x- and y-axes are set arbitrarily due to the absence of shear force measurements at these points. In addition, we integrate the real-time pose of each sensing point calculated using the robot’s forward kinematics into the node feature representation.
\subsubsection{Edge Feature}
The edge features are defined by the kinematic distances between the connected nodes. An alternative feature, which we did not adopt, is the relative pose between the frames of two connected nodes, offering richer geometric information. However, for graphs with thousands of nodes, computing relative poses for all edges is computationally expensive and fails to meet the real-time constraints of tactile gesture recognition. Future work could explore parallel computing and optimization techniques to accelerate this process, enabling more accurate and enriched geometric representations for improved recognition performance.
\subsubsection{Network Structure}
The neural network structure is illustrated in Fig. \ref{fig_15}b. Given the large number of sensing points on the robot’s skin, establishing edges between all sensors incurs a significant computational cost. To mitigate this, we first apply a K-Nearest Neighbors (KNN) layer to dynamically construct edges in real time based on kinematic distances, connecting each node only to its $k$ nearest neighbors.
To incorporate temporal information, we adopt a window-based processing approach. A First-In, First-Out (FIFO) buffer stores all graph data within a 2-second window, as most tactile gestures are typically completed within this duration. We then used a GNN based on the Equivariant Graph Convolutional Layer (EGCL) \cite{satorras2021n} to process the entire batch of graphs. This equivariant layer effectively preserves and exploits geometric information in $\mathrm{SE}(3)$ and has demonstrated efficiency in leveraging scarce data in robot applications, thereby enhancing training performance \cite{zhu2022sample, doi:10.1126/scirobotics.adn4008, huorbitgrasp}.
Our recognizer operates at 50Hz, meaning the FIFO buffer maintains a history of 100 dynamic graphs.
\subsubsection{Dynamics Connection}
\begin{figure}[!t]
\begin{subfigure}{.5\linewidth}
    \centering
    \includegraphics[height=.20\textheight]{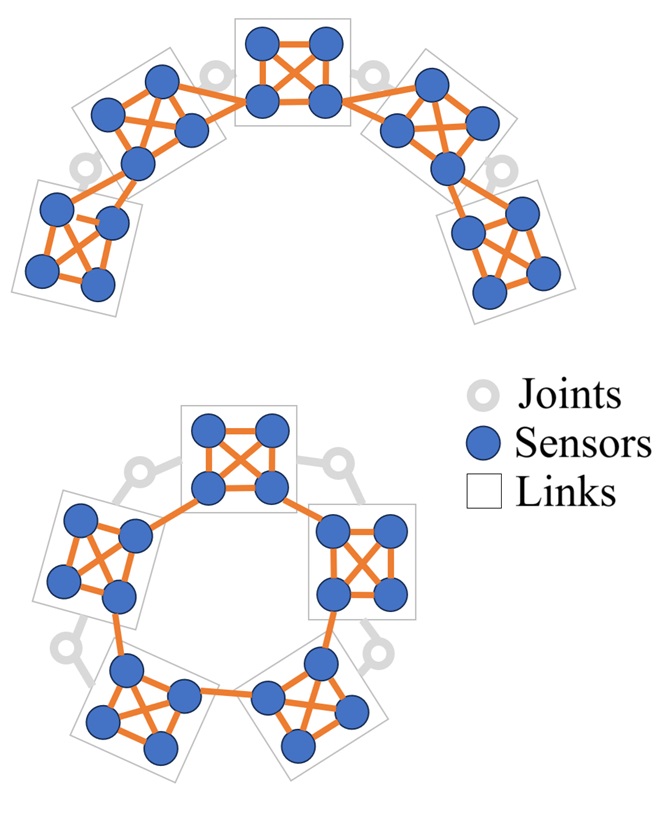}
    \caption{}
\end{subfigure}
\begin{subfigure}{.44\linewidth}
    \centering
    \includegraphics[height=.20\textheight]{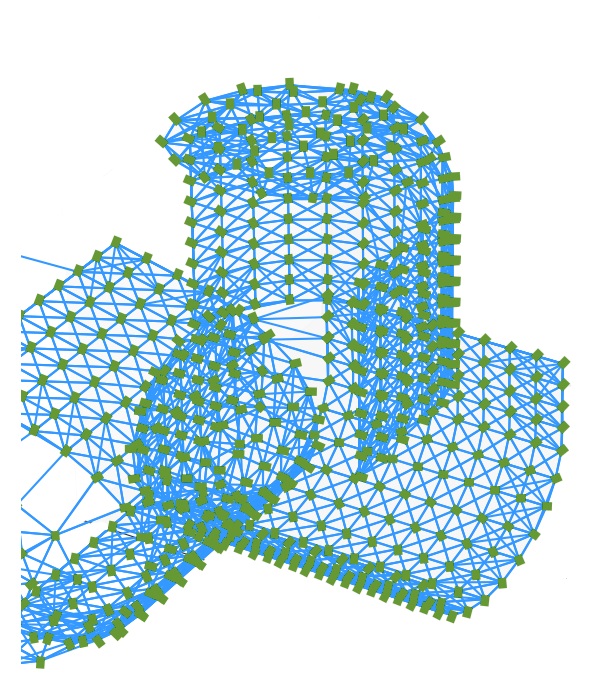}
    \caption{}
\end{subfigure}
 \caption{(a) Dynamic edge connection by real-time KNN. The geometry of the graph changes with the robot movement; (b) static edge connection of UR5 robot skin.}
 \label{fig_18}
\vspace{-10pt}
\end{figure}
\begin{figure}[!t]
 \centering
 \includegraphics[width=.8\linewidth]{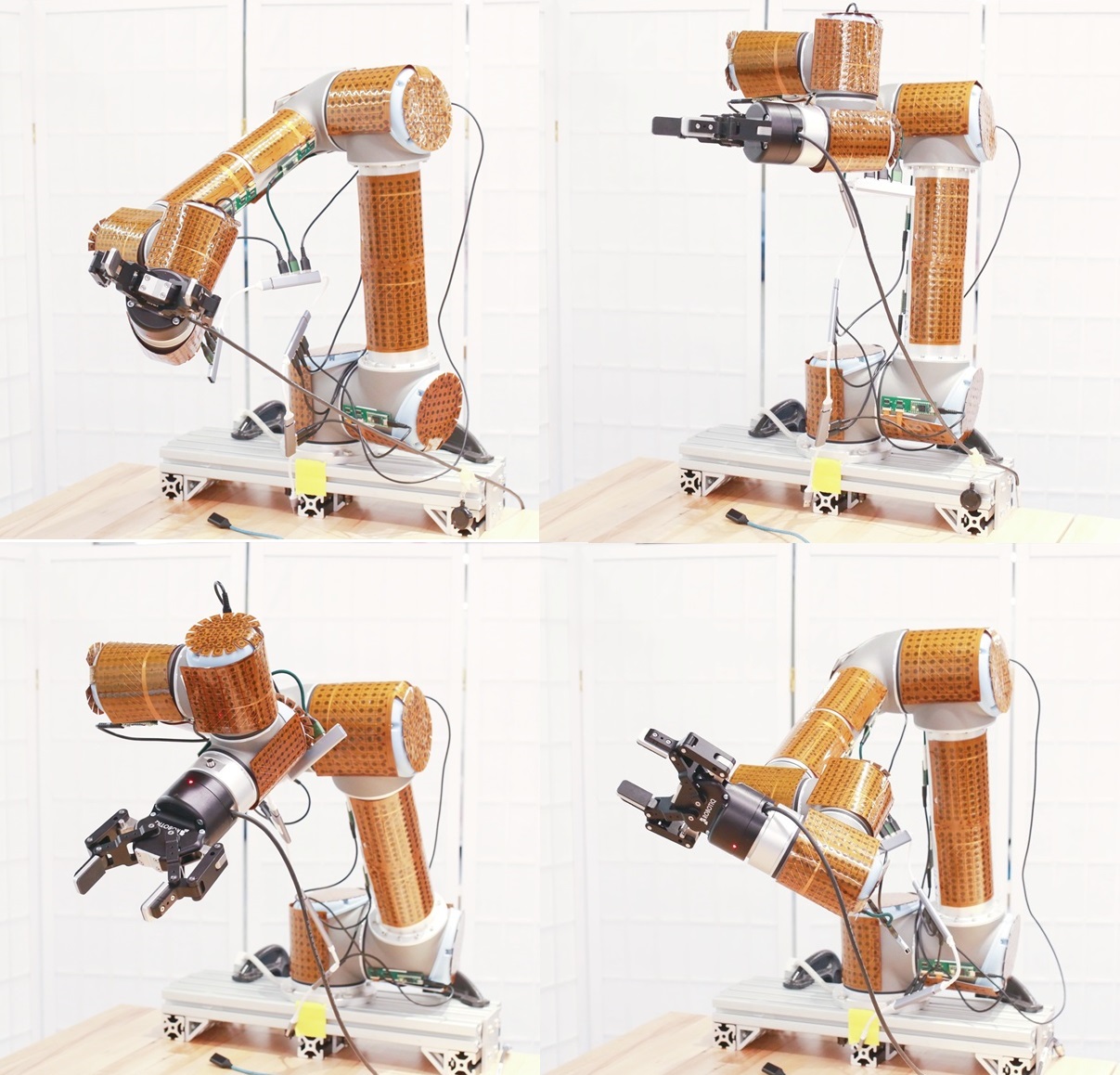}
 \caption{During data collection, the robot remains stationary in some random poses.}
 \label{fig_35}
\vspace{-10pt}
\end{figure}
\begin{figure*}[!t]
 \centering
\includegraphics[width=\linewidth]{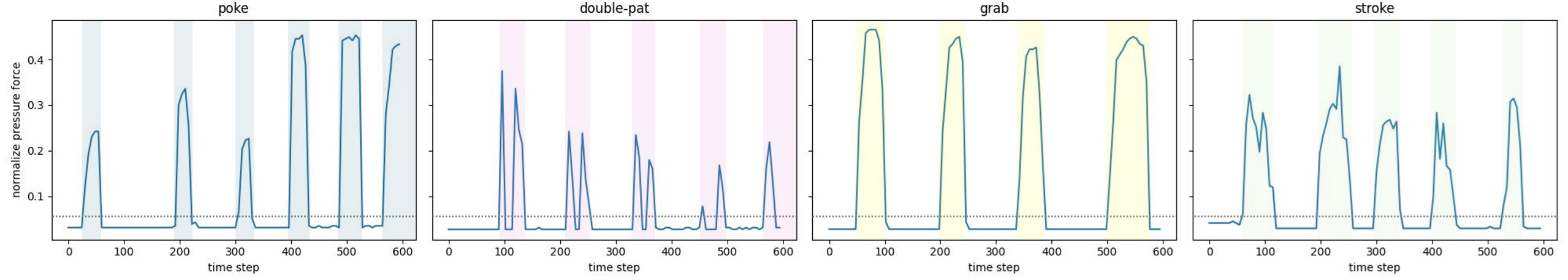}
 \caption{Threshold-based automatic data labeling.}
 \label{fig_5}
\vspace{-10pt}
\end{figure*}
To address the high computational complexity of generating dynamic connections using KNN in a continuously moving robot, we propose an optimized approach leveraging the static nature of certain connections. Despite the overall dynamic connectivity of the robot's nodes, we observe that the relative positions between sensing points within individual skin patches remain unchanged, regardless of the robot's movement. This observation allows us to pre-define these connections offline, significantly reducing the computation required during operation. The resulting graph, detailed in our results, contains 16,872 edges, as illustrated in Fig. \ref{fig_18}b. 

On the basis of pre-computed static connections, we have enhanced the graph's real-time responsiveness with dynamic connections. To effectively manage the computational demands of generating dynamic connections as the robot moves, we focus on analyzing those sensing points that are activated—specifically, those whose readings exceed a predetermined threshold. By applying the KNN algorithm only to these triggered sensing points and their adjacent nodes, we significantly reduce the scope of real-time computation. This method ensures efficient computation while maintaining accurate connectivity representation, crucial for the robot's operational integrity.
\subsection{Data Collection}
We collected a training dataset consisting of 581 distinct tactile gestures, which includes an even distribution of pat, poke, grab, and stroke gestures performed in various robot poses, as some are depicted in Fig. \ref{fig_35}. Additionally, we set aside 95 samples as the testing dataset. We recorded the real-time spatial pose and pressure readings from each touch point at a frequency of 50 Hz. During data collection, the robot remains stationary in each pose to ensure the stability of tactile signals. Typically, each gesture is executed in less than one second, with a two-second interval between consecutive gestures. A portion of the collected data is shown in Fig. \ref{fig_5}. In the figure, we present the maximum readings across all sensors to highlight distinct temporal features. Initially, we manually labeled the approximate range for each batch of tactile gestures. Within these ranges, we applied a threshold-based automatic data labeling method, incorporating a minimum window size to mitigate noise and prevent issues such as the separation of double-tap gestures.
\section{EXPERIMENTS}
\subsection{Hyperparameters}
\begin{figure}[!t]
 \centering
 \includegraphics[width=\linewidth]{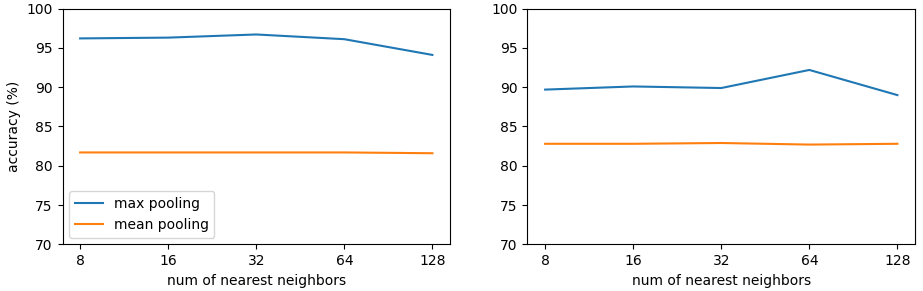}
 \caption{Classification accuracy: (left) on validation set; (right) on test set.}
 \label{fig_32}
\vspace{-10pt}
\end{figure}

We evaluated the performance of our models under various parameter configurations, as depicted in Fig. \ref{fig_32}. The graph on the left illustrates the validation accuracy, whereas the graph on the right depicts the test accuracy. Our analysis focused on the influence of different numbers of neighbors used in generating the KNN graph on model accuracy and the impact of the pooling layer within the GNN. We observed that varying the number of neighbors (K) did not significantly affect accuracy, allowing us to select a lower number of neighbors (K=8) for faster classification performance. However, the choice of pooling method proved to be consequential; max pooling demonstrated superior performance compared to mean pooling. This is likely due to the temporal and spatial sparsity of tactile signals, where most sensors record no readings most of the time. As a result, mean pooling tends to dilute the influence of active sensors, as the average is dominated by zeros, whereas max pooling effectively captures the peak values from those few activated sensors, enhancing the model's responsiveness to meaningful tactile interactions.
\begin{table}[t]
\scriptsize
    \centering
     \begin{tabular}{|c | c c c c c|} 
     \arrayrulecolor{gray}
     \hline
     \rowcolor{LightCyan}
      & Ours & CNN-LSTM \cite{sarwar2021large} & SVM \cite{lin2021event} & KNN & MLP  \\ 
     \hline\hline
     \rowcolor{light-gray}
      \cellcolor[rgb]{0.772549, 0.878431, 0.705882}Validation& 
      \textbf{96.1\%} & 
      95.7\% & 81.7\% & 90.3\% & 92.6\%\\
     \rowcolor{light-gray}
     \cellcolor[rgb]{0.772549, 0.878431, 0.705882}Test& 
      \textbf{91.1\%} & 
      80.4 \% & 82.8\% & 61.5\% & 76.1\%\\
     \hline
    \end{tabular}
    \caption{Recognition accuracy comparison}
    \label{tab_3}
    \vspace{-10pt}
\end{table}
\begin{figure*}[!t]
\begin{subfigure}{.17\linewidth}
    \centering
    \includegraphics[height=.15\textheight]{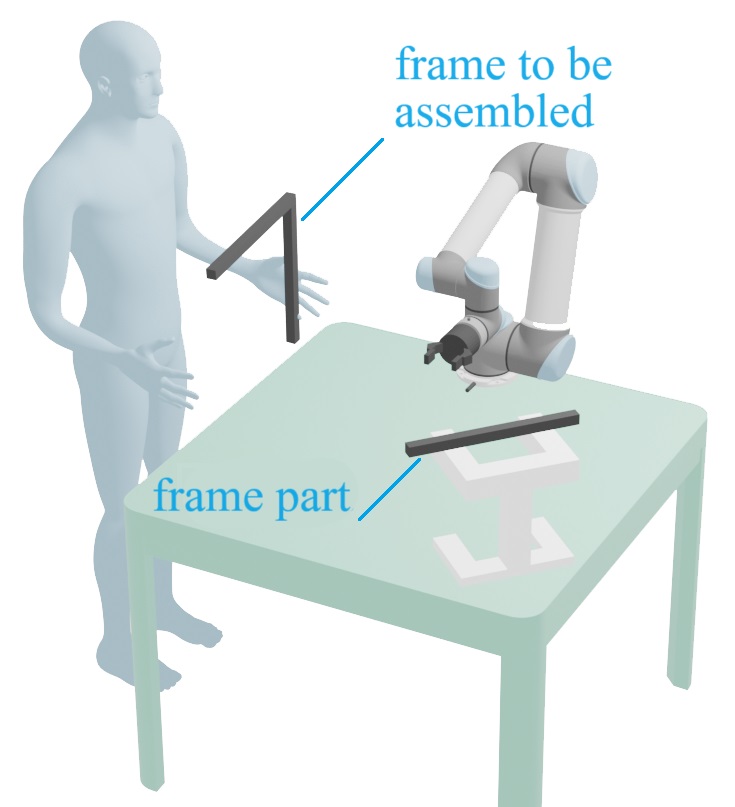}
    \caption{}
    \label{fig_36}
\end{subfigure}
\begin{subfigure}{.26\linewidth}
    \centering
    \includegraphics[height=.15\textheight]{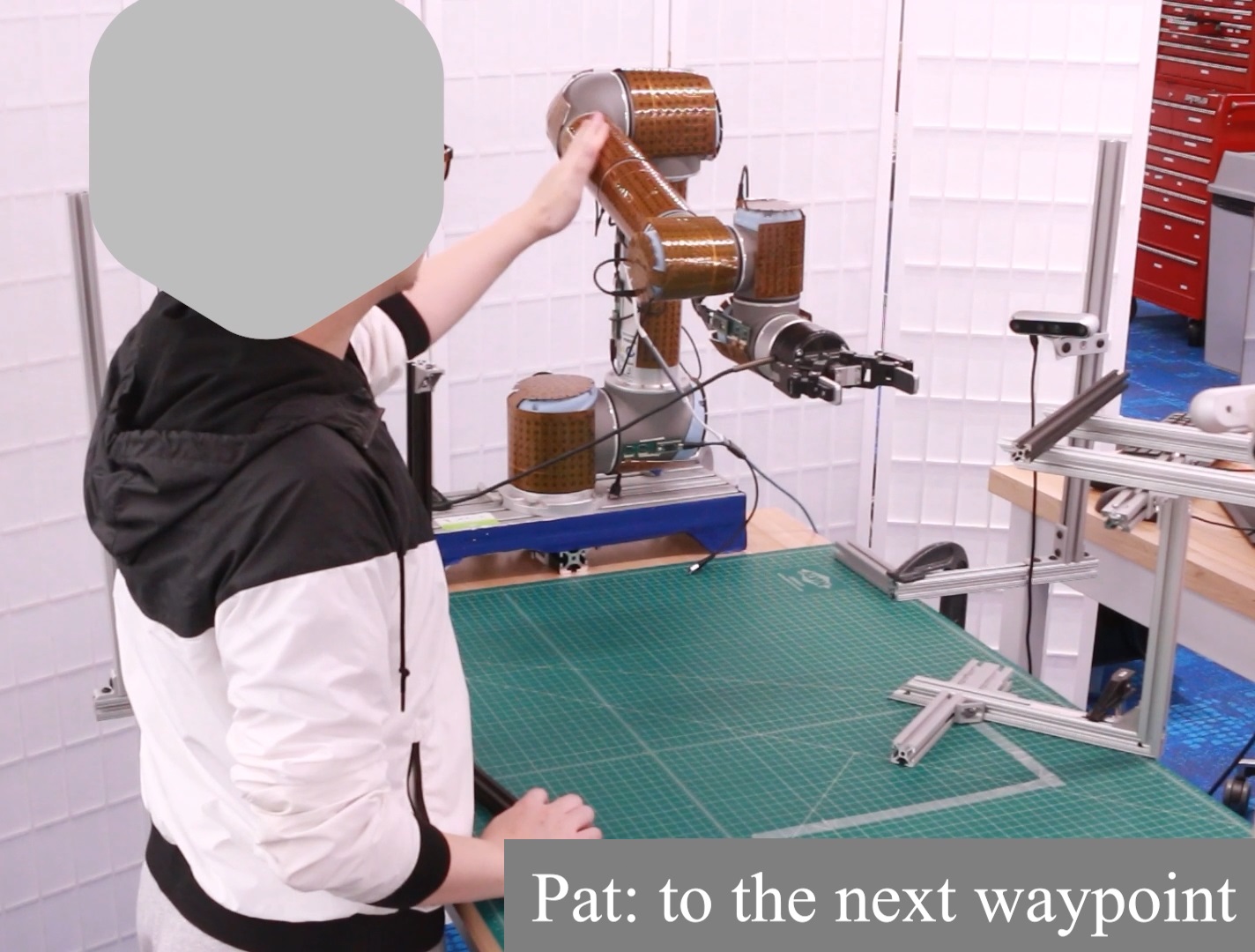}
    \caption{}
    \label{fig_202}
\end{subfigure}
\begin{subfigure}{.275\linewidth}
    \centering
    \includegraphics[height=.15\textheight]{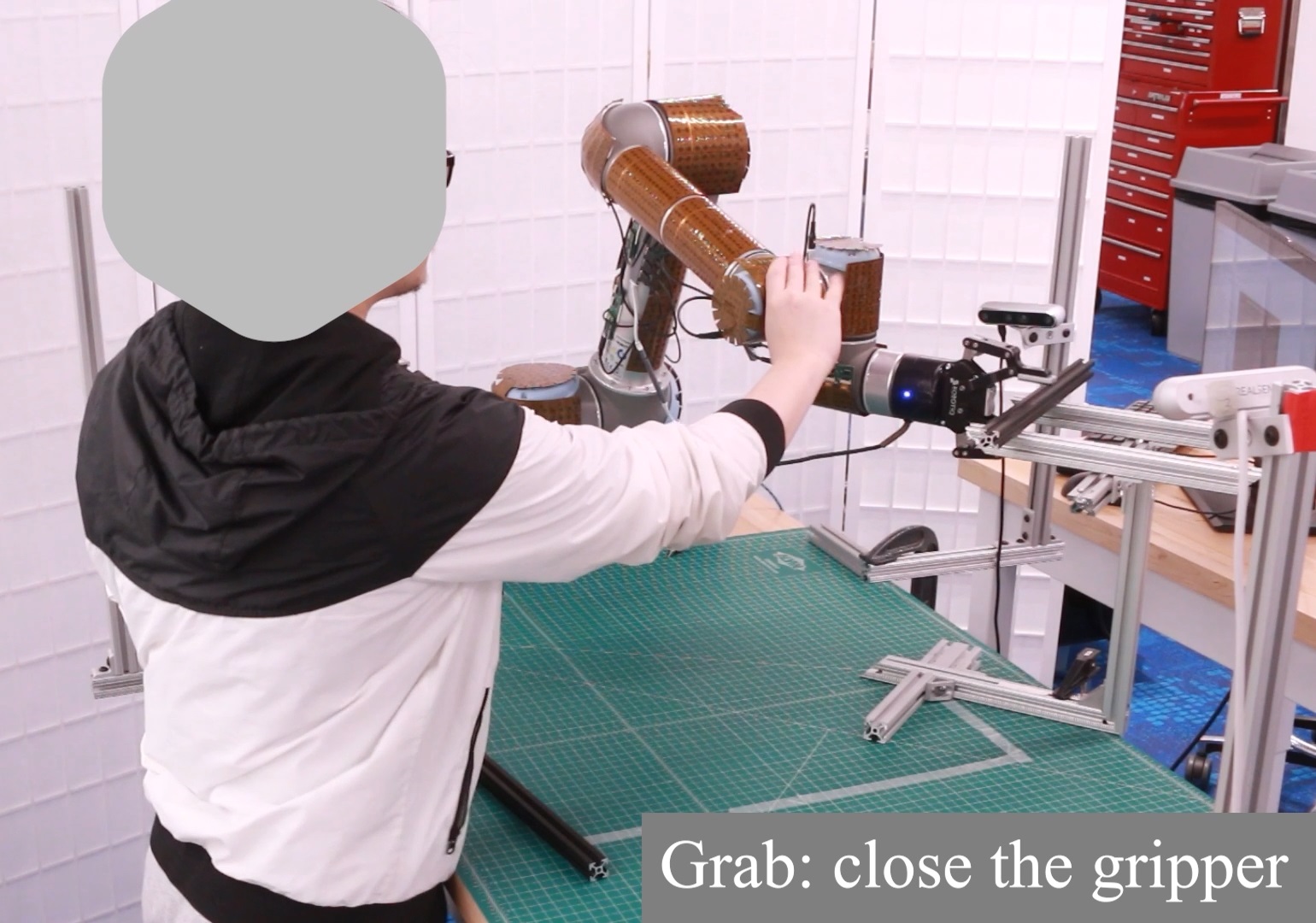}
    \caption{}
    \label{fig_203}
\end{subfigure}
\begin{subfigure}{.27\linewidth}
    \centering
    \includegraphics[height=.15\textheight]{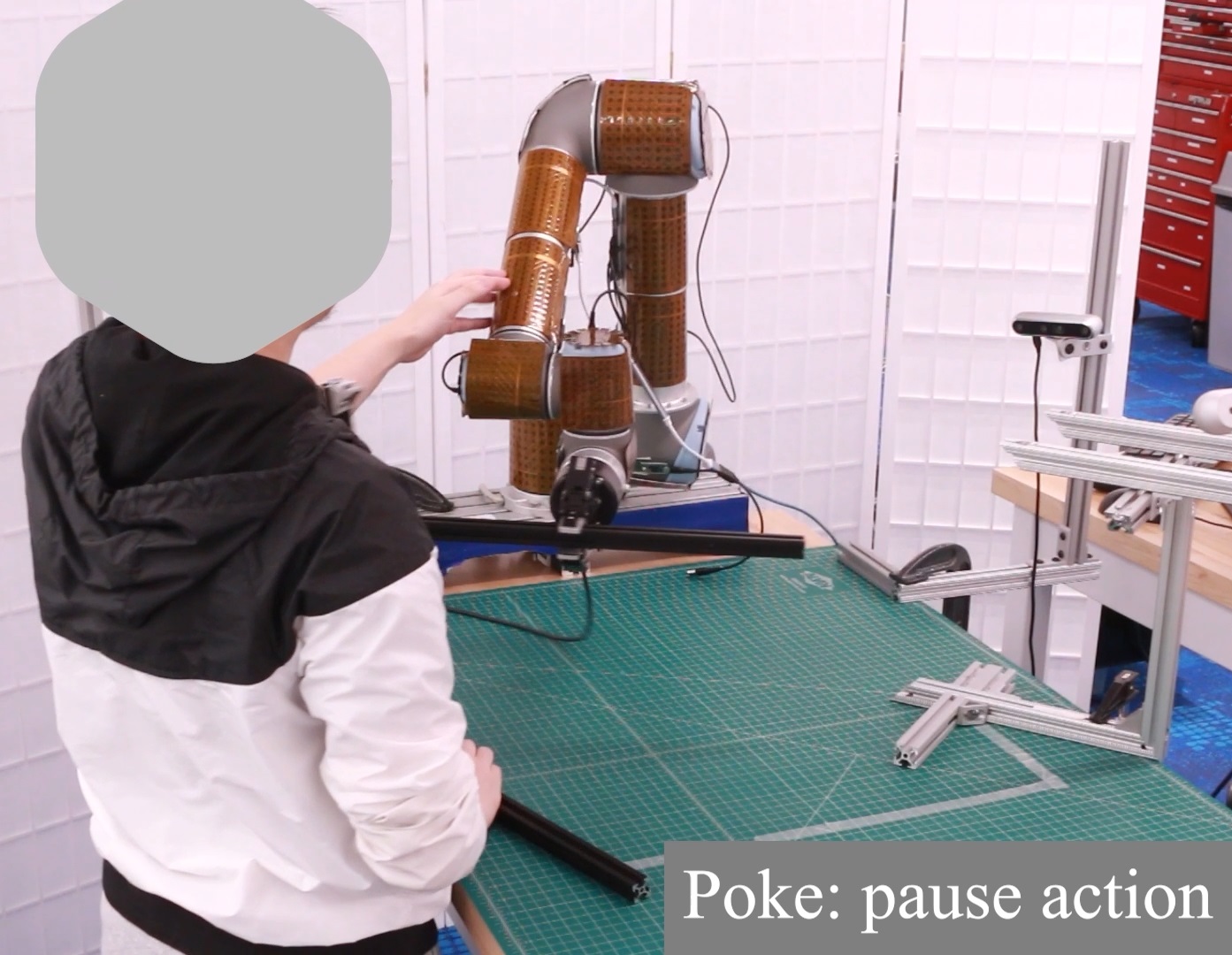}
    \caption{}
\end{subfigure}
 \caption{(a) Human-robot collaboration task; Mapping tactile gestures to robot actions: (b) double-pat: to the next waypoint; (c) grab: close the gripper; (d) poke: pause the robot.}
\end{figure*}
\subsection{Baselines Comparison}
We conducted a comparative analysis of various machine-learning-based tactile gesture recognition methods, including both deep learning and traditional non-deep-learning methods, using our dataset. The results, summarized in TABLE \ref{tab_3}, illustrate that our GNN-based solution significantly outperforms previous methods on both validation and test datasets. Specifically, in the CNN-LSTM method, each skin patch is processed as a separate channel for CNN input, where data from the same skin patch is reshaped into a rectangular grid with padding to emulate image-like data processing. In contrast, traditional approaches like SVM, KNN, and MLP simply flatten all sensor signals into a 2112-dimensional input vector. Our GNN approach distinguishes itself by integrating geometric information about the real-time poses of sensors, which enhances the network's ability to capture crucial spatial relationships.
\begin{table}[t]
    \centering
     \begin{tabular}{|c | c c c c c|} 
     \arrayrulecolor{gray}
     \hline
     \rowcolor{LightCyan}
      & Ours & CNN-LSTM \cite{sarwar2021large} & SVM \cite{lin2021event} & KNN & MLP  \\ 
     \hline\hline
     \rowcolor{light-gray}
      \cellcolor[rgb]{0.772549, 0.878431, 0.705882}Validation& 
      3.81 & 
       2.56& 737.67
 & 30.82
 & 2.14
\\
     \hline
    \end{tabular}
    \caption{One step runtime comparison (milliseconds)}
    \label{tab_2}
    \vspace{-10pt}
\end{table}

Due to the stringent real-time requirements of tactile gesture recognition, we also evaluated the runtime performance of various classifiers for single-step predictions, with results detailed in TABLE \ref{tab_2}. For traditional machine learning methods like SVM and KNN, we utilized the scikit-learn library \cite{scikit-learn}. Deep learning models were implemented using PyTorch. Our analysis shows that traditional methods, specifically SVM and KNN, exhibit longer prediction times compared to deep learning alternatives, rendering them less suitable for real-time applications where rapid processing is critical. This contrast underscores the efficiency of deep learning models in handling tasks requiring fast response times, aligning them more closely with the demands of real-time tactile gesture recognition.
\subsection{Real Robot Test}
To demonstrate the effectiveness of our tactile gesture recognition technology, we designed a simple human-robot collaboration experiment, as illustrated in the Fig. \ref{fig_36}. Human workers are tasked with assembling a metal frame, while a collaborative robot equipped with our E-skin is responsible for delivering frame part from the nearby rack to the workers. Through a series of mappings between tactile gestures and robot actions, workers are able to alter the robot's actions by employing different tactile gestures. Readers can refer to the video for a live demonstration.
\begin{itemize}
    \item \textbf{poke}: pause/resume the robot
    \item \textbf{double-pat}: robot moves to the next waypoint
    \item \textbf{grab}: close/release the gripper
\end{itemize}
The result suggests the human worker can successfully issue commands only by applying tactile gestures on the robot body regardless of the robot pose, which supports a new method of human-robot interaction.
\section{CONCLUSION}
In this work, we introduced our tactile gesture recognition system based on full-body electronic skin for robots. We used our developed E-skin with 2112 sensor points, adapted for an UR5 robot. We also developed an equivariant GNN-based recognizer to detect four types of tactile gestures: double-pat, poke, grab, and stroke. By incorporating the geometric information of the sensors into the GNN and using a dynamic KNN layer to infer the positional relationships of the sensors, our recognizer achieved a significant accuracy improvement compared to previous works. By mapping the tactile gestures to the robot's actions in a human-robot collaboration task, we achieved a workflow where the robot can be controlled through human tactile gesture interactions. Future work includes developing robot skin with higher precision and more modalities, adapting it to various types of robots, and enriching the library of tactile gestures and robot actions.
\addtolength{\textheight}{-0cm}   




\bibliographystyle{IEEEtran}
\bibliography{IEEEabrv,mybibfile}

\begin{thebibliography}{10}
\providecommand{\url}[1]{#1}
\csname url@rmstyle\endcsname
\providecommand{\newblock}{\relax}
\providecommand{\bibinfo}[2]{#2}
\providecommand\BIBentrySTDinterwordspacing{\spaceskip=0pt\relax}
\providecommand\BIBentryALTinterwordstretchfactor{4}
\providecommand\BIBentryALTinterwordspacing{\spaceskip=\fontdimen2\font plus
\BIBentryALTinterwordstretchfactor\fontdimen3\font minus \fontdimen4\font\relax}
\providecommand\BIBforeignlanguage[2]{{%
\expandafter\ifx\csname l@#1\endcsname\relax
\typeout{** WARNING: IEEEtran.bst: No hyphenation pattern has been}%
\typeout{** loaded for the language `#1'. Using the pattern for}%
\typeout{** the default language instead.}%
\else
\language=\csname l@#1\endcsname
\fi
#2}}

\bibitem{yohanan2012role}
S.~Yohanan and K.~E. MacLean, ``The role of affective touch in human-robot interaction: Human intent and expectations in touching the haptic creature,'' \emph{International Journal of Social Robotics}, vol.~4, pp. 163--180, 2012.

\bibitem{rew2000friends}
L.~Rew, ``Friends and pets as companions: Strategies for coping with loneliness among homeless youth,'' \emph{Journal of child and adolescent psychiatric nursing}, vol.~13, no.~3, pp. 125--132, 2000.

\bibitem{vormbrock1988cardiovascular}
J.~K. Vormbrock and J.~M. Grossberg, ``Cardiovascular effects of human-pet dog interactions,'' \emph{Journal of behavioral medicine}, vol.~11, pp. 509--517, 1988.

\bibitem{lin2021event}
S.~Lin, J.~Su, S.~Song, and J.~Zhang, ``An event-triggered low-cost tactile perception system for social robot’s whole body interaction,'' \emph{IEEE Access}, vol.~9, pp. 80\,986--80\,995, 2021.

\bibitem{choi2022deep}
H.~Choi, D.~Brouwer, M.~A. Lin, K.~T. Yoshida, C.~Rognon, B.~Stephens-Fripp, A.~M. Okamura, and M.~R. Cutkosky, ``Deep learning classification of touch gestures using distributed normal and shear force,'' in \emph{2022 IEEE/RSJ International Conference on Intelligent Robots and Systems (IROS)}.\hskip 1em plus 0.5em minus 0.4em\relax IEEE, 2022, pp. 3659--3665.

\bibitem{sarwar2021large}
M.~S. Sarwar and K.~Yamane, ``Large-area conformable sensor for proximity, light touch, and pressure-based gesture recognition,'' in \emph{2021 IEEE/RSJ International Conference on Intelligent Robots and Systems (IROS)}.\hskip 1em plus 0.5em minus 0.4em\relax IEEE, 2021, pp. 2700--2707.

\bibitem{jung2017automatic}
M.~M. Jung, M.~Poel, R.~Poppe, and D.~K. Heylen, ``Automatic recognition of touch gestures in the corpus of social touch,'' \emph{Journal on multimodal user interfaces}, vol.~11, pp. 81--96, 2017.

\bibitem{kong2022bioinspired}
D.~Kong, G.~Yang, G.~Pang, Z.~Ye, H.~Lv, Z.~Yu, F.~Wang, X.~V. Wang, K.~Xu, and H.~Yang, ``Bioinspired co-design of tactile sensor and deep learning algorithm for human--robot interaction,'' \emph{Advanced Intelligent Systems}, vol.~4, no.~6, p. 2200050, 2022.

\bibitem{sun2017categories}
J.~Sun, E.~Billing, F.~Seoane, B.~Zhou, D.~H{\"o}gberg, and P.~Hemeren, ``Categories of touch: Classifying human touch using a soft tactile sensor,'' in \emph{The robotic sense of touch: From sensing to understanding, workshop at the IEEE International Conference on Robotics and Automation (ICRA), Singapore, May 29, 2017}, 2017.

\bibitem{alonso2017detecting}
F.~Alonso-Mart{\'\i}n, J.~J. Gamboa-Montero, J.~C. Castillo, {\'A}.~Castro-Gonz{\'a}lez, and M.~{\'A}. Salichs, ``Detecting and classifying human touches in a social robot through acoustic sensing and machine learning,'' \emph{Sensors}, vol.~17, no.~5, p. 1138, 2017.

\bibitem{zhan2023enable}
L.~Zhan, Y.~Cao, Q.~Chen, H.~Guo, J.~Gao, Y.~Luo, S.~Guo, G.~Zhou, and J.~Gong, ``Enable natural tactile interaction for robot dog based on large-format distributed flexible pressure sensors,'' \emph{arXiv preprint arXiv:2303.07595}, 2023.

\bibitem{wang2021organization}
P.~Wang, J.~Liu, F.~Hou, D.~Chen, Z.~Xia, and S.~Guo, ``Organization and understanding of a tactile information dataset tacact for physical human-robot interaction,'' in \emph{2021 IEEE/RSJ International Conference on Intelligent Robots and Systems (IROS)}.\hskip 1em plus 0.5em minus 0.4em\relax IEEE, 2021, pp. 7328--7333.

\bibitem{kubus2017robust}
D.~Kubus, A.~Muxfeldt, K.~Kissener, J.~Haus, and J.~Steil, ``Robust recognition of tactile gestures for intuitive robot programming and control,'' in \emph{2017 IEEE/RSJ International Conference on Intelligent Robots and Systems (IROS)}.\hskip 1em plus 0.5em minus 0.4em\relax IEEE, 2017, pp. 1643--1650.

\bibitem{zhou2017textile}
B.~Zhou, C.~A. Velez~Altamirano, H.~Cruz~Zurian, S.~R. Atefi, E.~Billing, F.~Seoane~Martinez, and P.~Lukowicz, ``Textile pressure mapping sensor for emotional touch detection in human-robot interaction,'' \emph{Sensors}, vol.~17, no.~11, p. 2585, 2017.

\bibitem{bronstein2021geometric}
M.~M. Bronstein, J.~Bruna, T.~Cohen, and P.~Veli{\v{c}}kovi{\'c}, ``Geometric deep learning: Grids, groups, graphs, geodesics, and gauges,'' \emph{arXiv preprint arXiv:2104.13478}, 2021.

\bibitem{albini2020pressure}
A.~Albini and G.~Cannata, ``Pressure distribution classification and segmentation of human hands in contact with the robot body,'' \emph{The International Journal of Robotics Research}, vol.~39, no.~6, pp. 668--687, 2020.

\bibitem{albini2018tactile}
------, ``Tactile images generation from contacts involving adjacent robot links,'' in \emph{2018 27th IEEE International Symposium on Robot and Human Interactive Communication (RO-MAN)}.\hskip 1em plus 0.5em minus 0.4em\relax IEEE, 2018, pp. 306--312.

\bibitem{garcia2019tactilegcn}
A.~Garcia-Garcia, B.~S. Zapata-Impata, S.~Orts-Escolano, P.~Gil, and J.~Garcia-Rodriguez, ``Tactilegcn: A graph convolutional network for predicting grasp stability with tactile sensors,'' in \emph{2019 International Joint Conference on Neural Networks (IJCNN)}.\hskip 1em plus 0.5em minus 0.4em\relax IEEE, 2019, pp. 1--8.

\bibitem{funabashi2022multi}
S.~Funabashi, T.~Isobe, F.~Hongyi, A.~Hiramoto, A.~Schmitz, S.~Sugano, and T.~Ogata, ``Multi-fingered in-hand manipulation with various object properties using graph convolutional networks and distributed tactile sensors,'' \emph{IEEE Robotics and Automation Letters}, vol.~7, no.~2, pp. 2102--2109, 2022.

\bibitem{yang2023tacgnn}
L.~Yang, B.~Huang, Q.~Li, Y.-Y. Tsai, W.~W. Lee, C.~Song, and J.~Pan, ``Tacgnn: Learning tactile-based in-hand manipulation with a blind robot using hierarchical graph neural network,'' \emph{IEEE Robotics and Automation Letters}, vol.~8, no.~6, pp. 3605--3612, 2023.

\bibitem{fan2022graph}
W.~Fan, H.~Bo, Y.~Lin, Y.~Xing, W.~Liu, N.~Lepora, and D.~Zhang, ``Graph neural networks for interpretable tactile sensing,'' in \emph{2022 27th International Conference on Automation and Computing (ICAC)}.\hskip 1em plus 0.5em minus 0.4em\relax IEEE, 2022, pp. 1--6.

\bibitem{kulkarni2024tactile}
S.~Kulkarni, S.~Funabashi, A.~Schmitz, T.~Ogata, and S.~Sugano, ``Tactile object property recognition using geometrical graph edge features and multi-thread graph convolutional network,'' \emph{IEEE Robotics and Automation Letters}, 2024.

\bibitem{doi:10.1126/scirobotics.adn4008}
\BIBentryALTinterwordspacing
M.~Iskandar, A.~Albu-Schäffer, and A.~Dietrich, ``Intrinsic sense of touch for intuitive physical human-robot interaction,'' \emph{Science Robotics}, vol.~9, no.~93, p. eadn4008, 2024. [Online]. Available: \url{https://www.science.org/doi/abs/10.1126/scirobotics.adn4008}
\BIBentrySTDinterwordspacing

\bibitem{zhu2022sample}
X.~Zhu, D.~Wang, O.~Biza, G.~Su, R.~Walters, and R.~Platt, ``Sample efficient grasp learning using equivariant models,'' \emph{arXiv preprint arXiv:2202.09468}, 2022.

\bibitem{satorras2021n}
V.~G. Satorras, E.~Hoogeboom, and M.~Welling, ``E (n) equivariant graph neural networks,'' in \emph{International conference on machine learning}.\hskip 1em plus 0.5em minus 0.4em\relax PMLR, 2021, pp. 9323--9332.

\bibitem{huorbitgrasp}
B.~Hu, X.~Zhu, D.~Wang, Z.~Dong, H.~Huang, C.~Wang, R.~Walters, and R.~Platt, ``Orbitgrasp: {SE}(3)-equivariant grasp learning,'' in \emph{8th Annual Conference on Robot Learning}, 2024.

\bibitem{scikit-learn}
F.~Pedregosa, G.~Varoquaux, A.~Gramfort, V.~Michel, B.~Thirion, O.~Grisel, M.~Blondel, P.~Prettenhofer, R.~Weiss, V.~Dubourg, J.~Vanderplas, A.~Passos, D.~Cournapeau, M.~Brucher, M.~Perrot, and E.~Duchesnay, ``Scikit-learn: Machine learning in {P}ython,'' \emph{Journal of Machine Learning Research}, vol.~12, pp. 2825--2830, 2011.

\end{thebibliography}

\end{document}